\documentclass[
]{ceurart}

\usepackage{listings}
\usepackage{subcaption}
\usepackage[colorinlistoftodos]{todonotes}
\begin{document}

\copyrightyear{2022}
\copyrightclause{Copyright for this paper by its authors.
  Use permitted under Creative Commons License Attribution 4.0
  International (CC BY 4.0).}

\conference{11th Italian Workshop on Artificial Intelligence and Robotics (AIRO 2024)}

\title{Real-Time Multimodal Signal Processing for HRI in RoboCup: Understanding a Human Referee}


\author[1]{Filippo Ansalone}[%
orcid=0009-0002-0492-4748,
email=ansalone.1950936@studenti.uniroma1.it,
]
\cormark[1]
\fnmark[1]

\author[1]{Flavio Maiorana}[%
orcid=0009-0003-2059-7254,
email=maiorana.2051396@studenti.uniroma1.it,
]
\cormark[1]
\fnmark[1]

\author[1]{Daniele Affinita}[%
orcid=0009-0000-9347-9847,
email=affinita.1885790@studenti.uniroma1.it,
]
\cormark[1]

\author[1]{Flavio Volpi}[%
orcid=0009-0004-9822-5124,
email=volpi.1884040@studenti.uniroma1.it,
]
\cormark[1]

\author[1]{Eugenio Bugli}[%
orcid=0009-0000-9540-681X,
email=bugli.1934824@studenti.uniroma1.it,
]
\cormark[1]

\author[1,2]{Francesco Petri}[%
orcid=0009-0008-6208-1498,
email=francesco.petri@uniroma1.it,
]
\cormark[1]

\author[1]{Michele Brienza}[
orcid=0009-0000-1549-9500,
email=brienza@diag.uniroma1.it,
]\cormark[1]

\author[1]{Valerio Spagnoli}[%
orcid=0009-0008-0284-9602,
email=spagnoli.1887715@studenti.uniroma1.it,
]
\cormark[1]

\author[3]{Vincenzo Suriani}[%
orcid=0000-0003-1199-8358,
email=vincenzo.suriani@unibas.it,
]
\cormark[1]

\author[1]{Daniele Nardi}[%
orcid=0000-0001-6606-200X,
email=nardi@diag.uniroma1.it,
]

\author[4]{Domenico D. Bloisi}[%
orcid=0000-0003-0339-8651,
email=domenico.bloisi@unint.eu,
]

\address[1]{Sapienza University of Rome}
\address[2]{Institute for Cognitive Sciences and Technologies, National Research Council, Italy} 
\address[3]{University of Basilicata}
\address[4]{University of International Studies of Rome – UNINT}

\cortext[1]{Corresponding author.}
\fntext[1]{These authors contributed equally.}

\begin{abstract}
Advancing human-robot communication is crucial for autonomous systems operating in dynamic environments, where accurate real-time interpretation of human signals is essential. RoboCup provides a compelling scenario for testing these capabilities, requiring robots to understand referee gestures and whistle with minimal network reliance. Using the NAO robot platform, this study implements a two-stage pipeline for gesture recognition through keypoint extraction and classification, alongside continuous convolutional neural networks (CCNNs) for efficient whistle detection. The proposed approach enhances real-time human-robot interaction in a competitive setting like RoboCup, offering some tools to advance the development of autonomous systems capable of cooperating with humans.

\end{abstract}


\maketitle

\section{Introduction}


Human-robot communication has evolved significantly, but it becomes challenging in competitive environments such as RoboCup, where robots must interpret human signals with high accuracy. In these settings, the challenge is to reduce the reliance on network-based communications in favor of multimodal signal processing. This shift aligns with the growing interest in developing robots capable of understanding human gestures and audio cues, such as referee signals during matches.
The challenge lies in the robots' ability to process and interpret these multimodal signals in real-time, despite the constraints of limited computational resources. In the context of RoboCup, where human referees convey critical game states and events through gestures and whistles, the need for precise and efficient recognition systems becomes evident. This paper explores the integration of multimodal perception of gestures and whistles using the NAO robot platform, focusing on achieving robust performance under real-time conditions while being compliant with the official competition rules.

We employ a two-stage pipeline approach for gesture recognition, combining keypoint extraction and classification to interpret referee poses accurately. Simultaneously, we utilize continuous convolutional kernel neural networks (CKCNNs) \cite{romero2022} for whistle detection, balancing accuracy with computational efficiency. The proposed methods demonstrate the potential for enhancing human-robot interaction in competitive environments, contributing to the ongoing development of robot synergy with humans.

\begin{figure}[t]
    \centering
    \begin{minipage}[t]{0.5\textwidth}
        \centering
        \includegraphics[width=\textwidth]{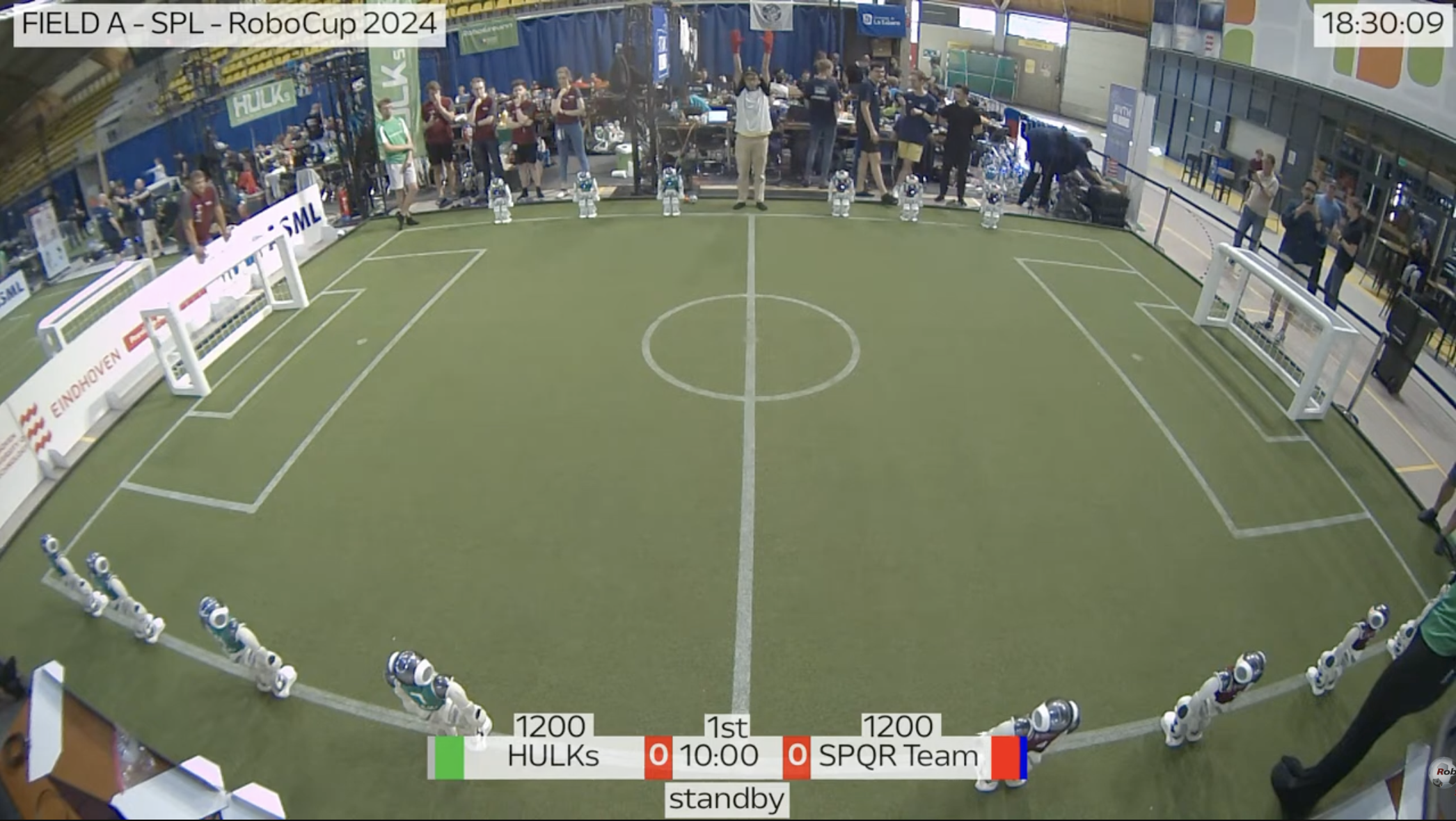}
    \end{minipage}%
    \begin{minipage}[t]{0.5\textwidth}
        \centering
        \includegraphics[width=\textwidth, height=0.181\textheight]{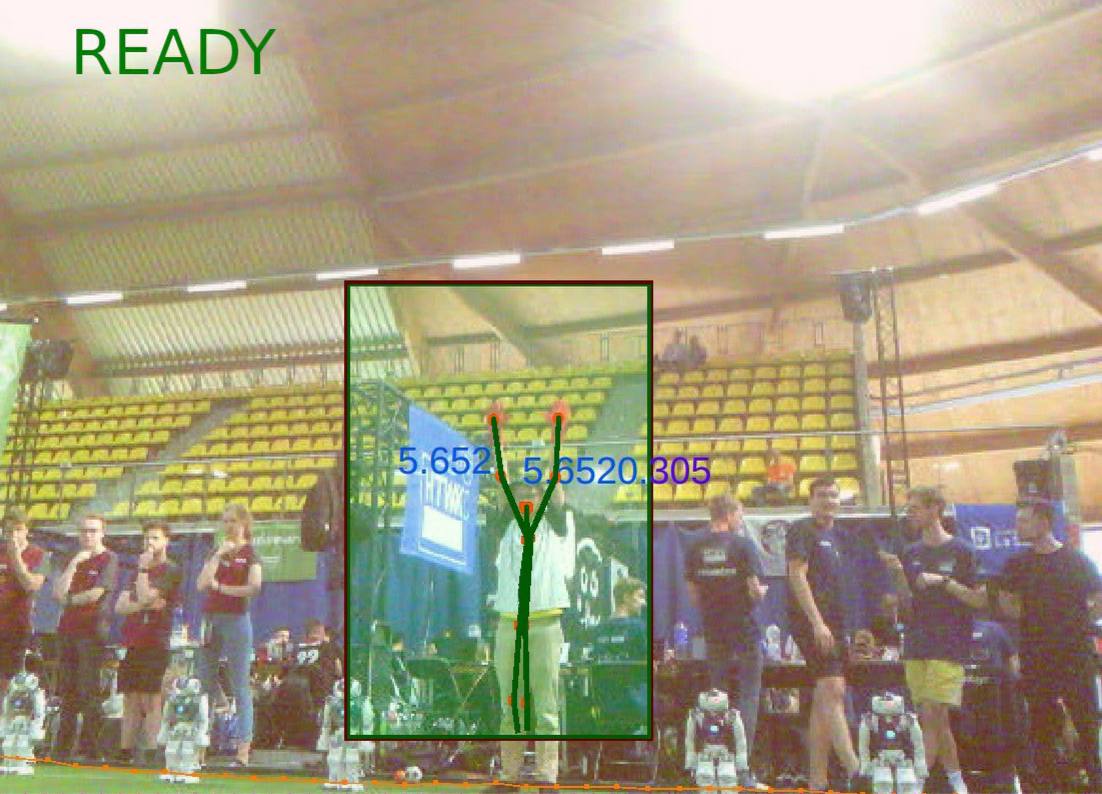}
    \end{minipage}
    \caption{Overview of the Robocup SPL field during the standby phase (left) and referee gesture detection from the robot's perspective (right). The right image highlights the region of interest (ROI) and displays the skeleton.}
    \label{fig:pose}
\end{figure}

\section{Related Work}

Interpreting human behavior has long been a central challenge in robotics. Humans communicate through various modalities, including vision, audio, and motion. This multimodal nature provides rich information that sensory input can capture and analyze. 

Recent advances in Deep Learning have facilitated the integration of multimodal data, significantly improving the comprehension of relationships within individual modalities, a key factor for precise message interpretation \cite{LIU20183} \cite{su2023recent}.

In the context of RoboCup, human-robot interaction is predominantly one-way, with human referees conveying game states and events to robots. A significant trend in the RoboCup SPL league is the progressive reduction of network communication in favor of human-like signal interpretation, allowing robots to interpret human signals more naturally.

In human soccer matches, gestures serve as a critical means of communication, especially in noisy environments such as stadiums. Previous works have extensively explored gesture recognition among agents using deep learning models, as seen in \cite{neto2019gesture}. A common approach is a two-stage pipeline, in which the person's skeleton is first extracted \cite{Kendall2015PoseNetAC} \cite{xiu2018pose}, followed by the classification of keypoint evolution over time. 
Specifically, Di Giambattista et al. \cite{10.1007/978-3-030-35699-6_28} employed OpenPose with Part Affinity Fields to extract the skeleton, using a subsequent network to analyze the relative positioning of keypoints for final pose prediction. Alternatively, single-stage pipelines \cite{yonet} \cite{cmc.2022.019586} offer end-to-end models, but require consideration of both spatial and temporal data from image sequences, often resulting in significantly larger models. Given that the NAO robot is an edge device with limited computational resources, we opted for a two-stage pipeline to maintain efficiency while ensuring accurate pose recognition.

Audio processing to detect specific sounds is an active research field, finding applications in various domains such as environmental monitoring \cite{white2022more}, security \cite{neri2022sound}, and sports analytics \cite{filippidis2019audio}. In particular, whistle detection has received attention in the context of sports, where referees’ whistles are used to signal important events during matches.
Unlike gestures, the whistling signal itself does not convey a specific meaning directly. Instead, it must be interpreted in the context of the current situation and game state, requiring a grounding \cite{jung2017affective} mechanism to relate the sound to relevant game events. A potential approach for whistle detection is to use LSTMs \cite{li2015lstm}, which offer the advantage of modeling long-range temporal dependencies and providing a larger context for analysis. However, they tend to be computationally expensive and slower due to their recurrent structure \cite{purwins2019deep}. Alternatively, computing the Fourier transform of the audio signal and using CNNs to process the resulting spectrogram is a more efficient solution \cite{naodevils}. Given that whistle recognition does not require modeling extensive temporal context, CNNs provide a better balance between accuracy and computational efficiency for our task.

\section{Methodology}

\begin{figure}[t]
    \centering
    \includegraphics[width=1.0\linewidth]{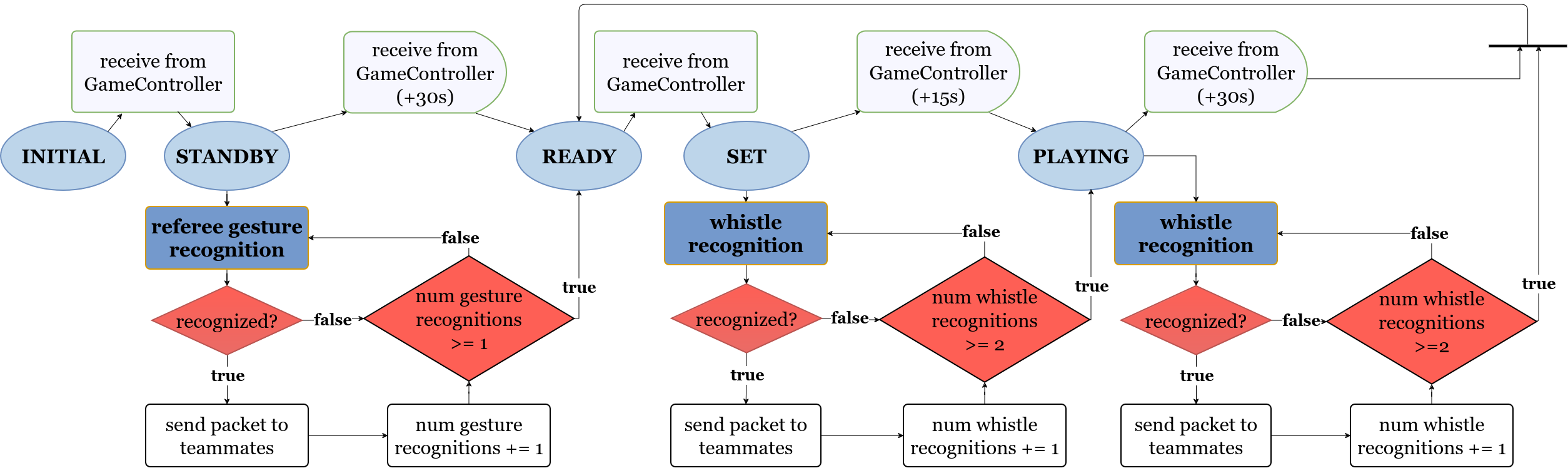}
    \caption{Triggered states during a game, for each robot in the field. It is important to highlight the integration of the modules that process referee's signals within the pipeline: a certain number of robots have to recognize a specific referee's signal (gesture for 4 consecutive camera frames or whistle) to move instantaneously to the following state, bypassing the delay associated to the message from Game Controller.}
    \label{fig:pipeline}
\end{figure}

To handle the detection of signals coming from the referee, a pipeline has been designed, involving many robots of the team. In Fig. \ref{fig:pipeline}, the pipeline is shown with the game states and the teammates.

\subsection{Whistle Recognition}

For whistle recognition, we employed Continuous Kernel Convolutional Neural Networks, which extend classical CNNs by using a kernel \textbf{parametrized by a small neural network}. CNNs excel in efficiently learning functions over structured data, like images or audio, by leveraging translation equivariance, albeit with a fixed receptive field size. In contrast, continuous kernel convolutions adapt to varying input lengths and resolutions, offering several advantages in audio processing: \begin{itemize} 
    \item The same architecture accommodates different preprocessing techniques, such as varying sampling rates, window sizes, or feature extraction methods (e.g., STFT or MFCC). 
    \item The number of parameters of the network is decoupled from its receptive field, allowing to have a long-range kernel with a relatively small number of parameters. 
\end{itemize} 
In our application, the basic building block is the CKBlock:
\begin{verbatim}
    input -> BatchNorm -> CKConv -> GELU -> DropOut -> Linear -> GELU -> + -> output
      |__________________________________________________________________|
\end{verbatim}
The CKConv layer is the core of the architecture, since it contains the kernel generation and convolution operation. 
The convolution operation is defined as $(x * \psi)(t) = \sum\limits_{c=1}^{N_{in}} \sum\limits_{\tau=0}^{t} x_c(\tau)\cdot\psi_c(t-\tau)$, which means that the convolver is now viewed as a vector-valued continuous function
$\psi: \mathbb{R} \rightarrow \mathbb{R}^{N_{out} \times N_{in}}$,
parametrized with a small neural network $MLP^{\psi}$:
\begin{itemize}
    \item The input is a relative position $(t-\tau)$ of the convolvee
    \item The output is the value $\psi(t-\tau)$ of the convolutional kernel at that position
\end{itemize}
The main consequence of this is that the kernel is arbitrarily large.

The entire network is a sequence of 4 CKBlocks, with a final fully connected
part. More specifically, the convolutional layers have a hidden size of 32. The
convolutional kernels are structured as simple 3-layer MLPs with hidden size 16.
We chose as kernel size 31, since an overly large kernel would overfit the training
data, while a too small kernel would need a deeper network. Overall, we reached
a network size of 59.1k trainable parameters. 

\subsubsection{Data gathering}

\paragraph{Structure and preprocessing}
In addition to the task of classifying an audio sample as either whistle or no-whistle, a critical challenge in RoboCup games is ensuring accurate predictions in the presence of background noise, such as crowd sounds, robot movements, and other environmental sounds. Therefore, the dataset \cite{naodevils} is a collection of audio files collected both in lab conditions and during the actual matches, using the robots' microphones. Since, on average there are few whistles in a match, the result is a heavily unbalanced dataset, with a ratio of $10:1$ (60000 no-whistle samples, 6000 whistle samples). The dataset was manually cleaned, removing many samples where the only noise source was the robot walking, or where there was silence. Also, the labelling happened manually through the software Audacity by extracting the audio events, defined as start and end of the whistle, in text files. These were then associated with the corresponding audio samples using the library \textit{Librosa} \cite{librosa}.

\paragraph{Feature extraction}
To extract the features, we perform a frequency analysis of the audio signal using
short-time fourier transforms. The result is, for each audio, a series of vectors
of shape \verb|(1,NUMBER_FREQUENCIES)|, where each vector represents the
frequency amplitudes of a window. We extracted 1024 frames per window at 44100
Hz. This resulted in every data sample being a vector of shape \verb|(1,513)|.

\subsection{Gesture Recognition}
For the recognition of a referee pose, we propose a 2 step architecture based on a pretrained key point extractor and then a classification module.

Since one of the goals in RoboCup is to optimize as much as possible each algorithm to grant a fast real-time execution, we had to rely on MoveNet Lightning\cite{movenet} which is a deep learning architecture based on MobileNetV2 \cite{MobileNetV2} specifically developed for real-time applications which takes as input a 192x192 RGB image. We adapt the Nao camera frames, featuring a resolution 640x480, by scaling and padding to match the input shape. Due to the distance of the referee from the robots, the image scaling down leads to a detail loss on our region of interest (ROI) and the key point extractor does not recognize the pose correctly. To overcome this issue, we implemented a crop on the ROI containing the referee, and then resized and padded it to the desired input shape. This crop is also useful to prevent the MoveNet to focus on a different person which is standing at the border of the field which may cause false readings making the entire pipeline more robust. Figure \ref{fig:pose} illustrates an example of the ROI selection and the pose estimation network in action, estimating the referee's skeleton. 

After the key point extraction, we needed to extract a good feature because classifying directly on the raw key point coordinates would be a much harder problem, especially with a small dataset. To address this problem we decided to calculate the angles of the joints that are more useful for our task. So for both the left and right sides of the body, the algorithm computes the angles between:
\vspace{-0.2cm}
\begin{itemize}
    \item Hip - Shoulder - Elbow
    \item Shoulder - Elbow - Wrist
\end{itemize}
\vspace{-0.2cm}
This procedure eventually computes less features that are, on the other hand, much more representative of our problem. In general, given 3 points (A, B, C) the angle is computed as:
\[\theta=atan2(BC_y,BC_x)-atan2(BA_y,BA_x)\]
This feature is better not only because it is easier to interpret but also because it grants scale and rotation invariance which are very useful considering that both the dataset and the classifier architecture were small. These two properties, together with the intrinsic translation equivariance provided by the CNN architecture, contribute to a generally more robust pipeline.

When a robot sees the pose for at least 4 consecutive frames, the recognition is succesful and a packet is sent to the team so that every robot can enter the ready state, as shown in Fig. \ref{fig:pipeline}.

\subsubsection{Data gathering}
The released rule that has to be followed states that \emph{``To announce the transition from standby to ready state, the referee will raise both hands over their head''}.

To this end, the dataset was collected by our team in a private environment, allowing for consistent conditions throughout the data acquisition process. This approach facilitated data gathering, which was subsequently manually labeled to ensure a high quality labeling.

\section{Results}
We evaluated separately the whistle and the gesture subsystems. 
Table \ref{tab:res} shows the results of the models on the test data and the real
scenario. In the whistle test data case, we reached a lower precision, due to the highly
imbalanced dataset. Whereas, in a real scenario, the detector worked pretty well. Lowers
precision could be a problem in cases of similar sounds to whistles that could
cause false detections. This can be easily mitigated by using a consensus
approach. In case of the real scenario, the distinction between playing and not
playing is made to show the difference between these two cases. When the robots
are playing, the whistle always comes after a goal is scored, and usually the
crowd cheers in such a situation. Therefore, especially when referees do not
whistle loudly, the model is not able to distinguish the whistle sound from the
crowd noise. On the other hand, when the robots are not playing, it means they
are waiting for a kick-off. In this case, there is usually less noise, and the
model is able to detect the whistles with high accuracy. The same pattern occurs in the gesture recognition case, in which high precision was preferred over recall to avoid incurring rule penalties.

\begin{table}[t]
\centering
\subfloat[Whistle Recognition Results based on 147440 test samples (frequency windows) and 73 real situations over 8 games]{
    \scalebox{0.82}{
        \begin{tabular}{ |c|c|c|c| } 
            \hline
             & Accuracy & Precision & Recall \\
            \hline
            Test & 98.02\% & 79.36\% & 90.35\% \\ 
            Real (Play) & 75\% & 100\% & 80\% \\ 
            Real (Ready/Set) & 100\% & 100\% & 100\% \\
            \hline
        \end{tabular}
    }
}
\hspace{0.7cm}
\subfloat[Gesture Recognition Results based on 153 test samples and 18 real situations over 8 games]{
    \scalebox{0.82}{
        \begin{tabular}{ |c|c|c|c|c| } 
            \hline
             & Accuracy & Precision & Recall & F1-Score \\
            \hline
            Test & 99\% & 99\% & 99\% & 99\% \\ 
            Real & 50\% & 100\% & 50\% & 66\% \\ 
            \hline
        \end{tabular}
    }
}
\caption{Overall performance evaluation of both networks used to interpret the human referee, reporting metrics from both the dataset and real scenarios.}
\label{tab:res}
\end{table}
Both pipelines are fast enough to run on a NAO robot in about 0.8 ms (whistle) and 200 ms (gesture).



\section{Conclusions}

This paper presents an approach to detecting audiovisual signals from a human in the context of a robot soccer game in real-time.  Using a two-stage pipeline for gestures and a CCNN for whistles, we balanced computational efficiency with accuracy on the NAO robot platform.

Our results showed strong performance in whistle detection, while gesture recognition faced challenges in real-world conditions, particularly in noisy environments. Future work will focus on enhancing noise resilience and improving gesture recognition to better handle dynamic scenarios.


\begin{acknowledgments}

  This work has been carried out while Francesco Petri and Michele Brienza were enrolled in the Italian National Doctorate on Artificial Intelligence run by Sapienza University of Rome. 
  We also acknowledge partial financial support from PNRR MUR project PE0000013-FAIR.
\end{acknowledgments}

\bibliography{sample-ceur}

\end{document}